\documentclass[letterpaper, 10 pt, conference]{ieeeconf}  % Comment this line out if you need a4paper
\IEEEoverridecommandlockouts                              % This command is only needed if 
\usepackage{xcolor}
\usepackage{amssymb}
\usepackage{graphicx}
\usepackage{mathtools}
\usepackage{amsmath}
\usepackage{gensymb}
\usepackage{float}
\usepackage{multirow}
\usepackage{makecell}
\usepackage{mathtools}
\usepackage{hyperref}
\hypersetup{
    colorlinks=true,
    linkcolor=blue,
    filecolor=magenta,      
    urlcolor=blue,
    pdftitle={Overleaf Example},
    pdfpagemode=FullScreen,
    }

\usepackage{tabularx}
    \newcolumntype{L}{>{\raggedright\arraybackslash}X}
\usepackage[utf8]{inputenc}
\usepackage[english]{babel}
\usepackage{amsthm}

\usepackage{subcaption}

\newcommand{\no}{\noindent}

\title{\LARGE \bf
Sim-to-Real Strategy for Spatially Aware Robot Navigation in Uneven Outdoor Environments
}
\author{Kasun Weerakoon$^1$, Adarsh Jagan Sathyamoorthy$^1$, and Dinesh Manocha$^2$% <-this % stops a space
\thanks{This work was supported in part by ARO Grants W911NF1910069, W911NF2110026  and U.S. Army Grant No. W911NF2120076. We acknowledge the support of the Maryland Robotics Center.}% <-this % stops a space
\thanks{$^{1}$ Authors are with Dept. of Electrical and Computer Engineering, University of Maryland, College Park, MD, USA. {\tt\footnotesize kasunw@umd.edu, asathyam@umd.edu}}
\thanks{$^{2}$ Author is with Dept. of Computer Science, University of Maryland, College Park, MD, USA. {\tt\footnotesize dm@cs.umd.edu}}
}

\begin{document}

\maketitle
\thispagestyle{empty}
\pagestyle{empty}

%%%%%%%%%%%%%%%%%%%%%%%%%%%%%%%%%%%%%%%%%%%%%%%%%%%%%%%%%%%%%%%%%%%%%%%%%%%%%%%%
\begin{abstract}

Deep Reinforcement Learning (DRL) is hugely successful due to the availability of realistic simulated environments. However, performance degradation during simulation to real-world transfer still remains a challenging problem for the policies trained in simulated environments. To close this sim-to-real gap, we present a novel hybrid architecture that utilizes an intermediate output from a fully trained attention DRL policy as a navigation cost map for outdoor navigation. Our attention DRL network incorporates a robot-centric elevation map, IMU data, the robot's pose, previous actions and goal information as inputs to compute a navigation cost-map that highlights non-traversable regions. We compute least-cost waypoints on the cost map and utilize the Dynamic Window Approach (DWA) with velocity constraints on high cost regions to follow the waypoints in highly uneven outdoor environments. Our formulation generates dynamically feasible velocities along stable, traversable regions to reach the robot's goals. We observe an increase of 5\% in terms of success rate, 13.09\% of decrease in average robot vibration, and a 19.33\% reduction in average velocity compared to end-to-end DRL method and state-of-the-art methods in complex outdoor environments. We evaluate benefits of our method using a Clearpath Husky robot in both simulated and real-world uneven environments. 

%Video and a full technical report are available at \href{https://gamma.umd.edu/terp/}{gamma.umd.edu/terp}. 

\end{abstract}

\section{Introduction}

Mobile robots have been used widely in numerous outdoor navigation applications including delivery, exploration, rescue, construction and surveying \cite{rescue_robot,agriculture,badgr,construction}. Such applications expect robots to traverse autonomously on complex uneven terrains including hills, rocks, ramps, curbs etc. A robot should be able to plan trajectories along the optimally traversable regions to perform  safe and stable navigation on such terrains.

Traditional methods proposed for robot navigation on unstructured terrains have incorporated techniques such as classification models \cite{matos2019terrain}, semantic segmentation \cite{ganav_arxiv} and potential fields \cite{potential_field,Shimoda_potential} to represent terrain traversability. However, these techniques heavily depend on heuristics or human annotations. To learn terrain features from the robot's perspective, deep reinforcement learning (DRL) strategies have been employed for optimal policy training to map environmental features (observations) with robots' actions \cite{DRL_review}. In general, simulation environments are utilized to train such DRL policies to alleviate safety concerns and excessive time consumption in real robot deployment.

Simulation environments provide a fast and convenient framework to process diverse large, diverse sets of environment data for policy training. However, in general, DRL policies learned in simulators cannot be  deployed directly in real robots due to a lack of high-fidelity simulations (sensory noise, unmodeled dynamics, etc.) and physics. This discrepancy between the real world and the simulators is called the \textit{sim-to-real gap} in the literature.  

Classical strategies to minimize the sim-to-real gap in robot navigation literature include domain adaptation, inverse dynamics model learning and parameter tuning using real world training \cite{truong2021bi,wigness2018robot,chen2022should,rana2020multiplicative}. However, many of these works only consider flat or indoor terrains which are relatively less complex compared to uneven outdoor terrains \cite{gao2021sim2real}. Several DRL methods \cite{josef2020deep,zhang2018robot,Nguyen} have also been trained and tested in simulated uneven terrains using sensory inputs such as elevation maps,  robot orientation, depth images,  RGB images, and point-clouds. Hu et al. \cite{hu2021sim} achieve better navigation performance on synthetic uneven terrains using a sim-to-real pipeline that can capture abrupt changes in surface normals and elevations. However, all the aforementioned methods are tested only in simulated, synthetic, or controlled environments.

% \adarsh{Maybe mention their limitations here and how terp overcame them. So there is continuity with the next para.}

\begin{figure}[t]
      \centering
      \includegraphics[width=\columnwidth,height=7cm]{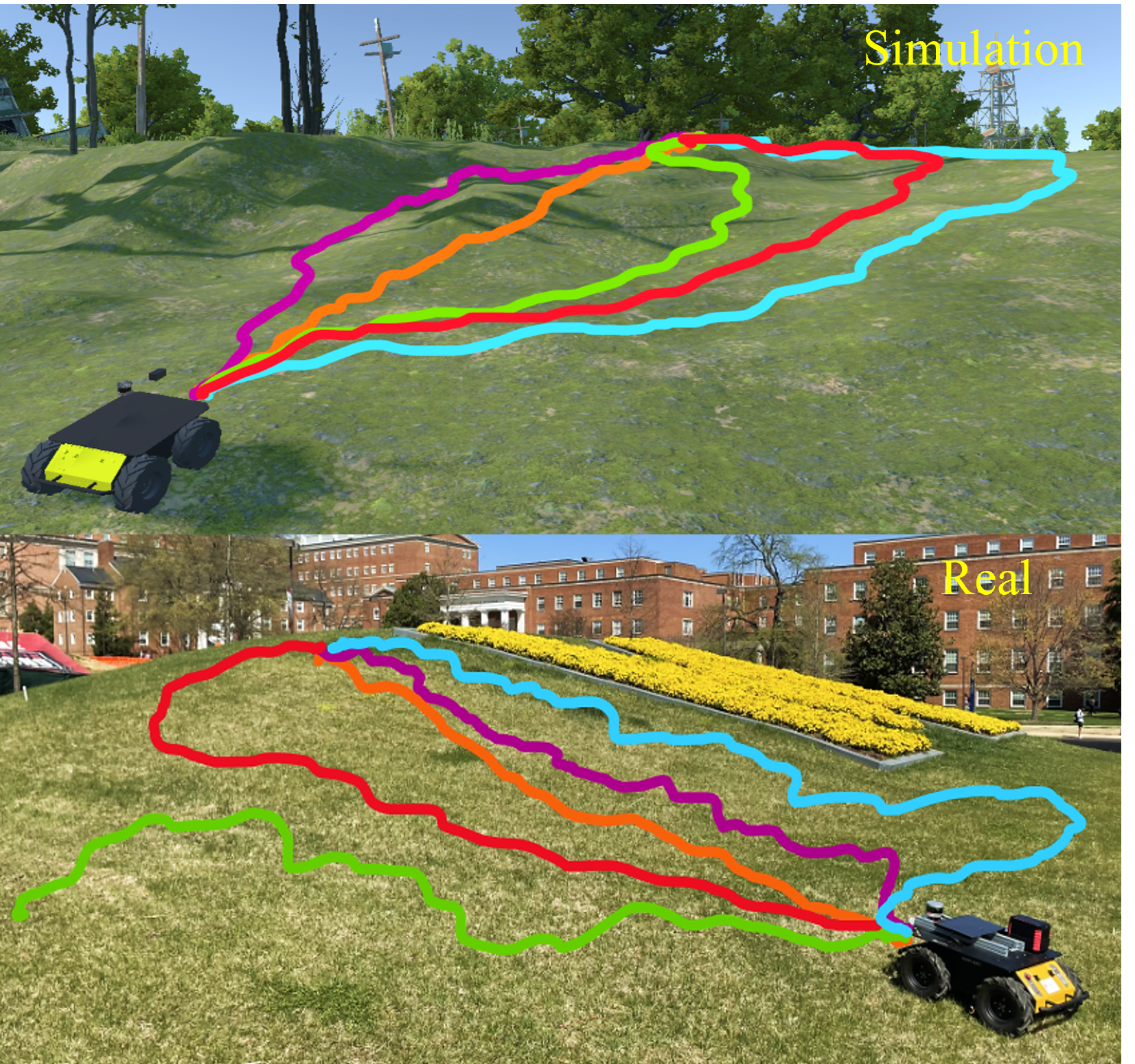}
      \caption {\small{Trajectories when navigating on simulated (top) and real-world (bottom) uneven outdoor terrains: Ours(red); Ours without attention(violet); End-to-end DRL (green); TERP(blue) \cite{terp_arxiv}; DWA(orange) \cite{DWA}. Our method generates trajectories along the least elevation gradients while controlling navigation velocities based on robot's current vibration and orientation to reduce the risk of robot flip-overs. Further, our formulation maintains comparable performance in both simulated and real world uneven outdoor environments, while the end-to-end method cannot.   }}
      \label{fig:cover-image}
      \vspace{-15pt}
\end{figure}

In our previous work TERP\cite{terp_arxiv}, we proposed a reliable planning method to generate trajectories avoiding reduced-stability regions in real uneven outdoor environments using DRL. Even though this approach performs well in both simulated and real world uneven terrains, it does not incorporate velocity control to avoid flip-overs on steep hills and rough terrains. The robot's stability not only depends on the trajectory but also on its velocity.

Main Contributions: We propose an extended version of our work TERP \cite{terp_arxiv} to perform stable and smooth robot navigation in uneven outdoor environments. The main contributions of our method include:

\begin{itemize}
    \item A sim-to-real strategy to utilize intermediate encoded features of a fully trained end-to-end DRL policy for perceiving regions that could cause instabilities for the robot during navigation in real world uneven outdoor environments.
    \item IMU and elevation gradient based rewards to identify critical elevations and rough surfaces in robot's vicinity. 
    \item Novel velocity space constraints on the Dynamic Window Approach (DWA)\cite{DWA} to penalize velocities that could cause robot flip-overs and high vibrations. Our formulation results in a 13.09\% decrease in average robot vibration, and a 19.33\% reduction in average velocity compared to end-to-end DRL methods and state-of-the-art methods in complex outdoor environments.
    % \adarsh{; This is unclear. We should expand this point more.}   
\end{itemize}

\section{Proposed Method} \label{sec:our-method}

In this section, we explain details of our method's perception and planning components: 1. Cost-map generation using a novel attention -based DRL network, 2. Waypoint calculation and 3. Navigation using DWA with a modified velocity space.
% \adarsh{is adaptation the right word?}. 
Our formulation uses 3D point-clouds, the robot's pose, IMU data and goal location as its sensory inputs. The point-cloud data is processed to obtain a robot-centric 2D elevation map. Our method's overall system architecture is presented in Fig. \ref{fig:system-architecture}.
% Further, IMU data is processed using PCA to obtain a low-dimensional representation.

\subsection{Attention DRL Network}

\subsubsection{Network Architecture}
We incorporate Deep Deterministic Policy Gradient (DDPG) \cite{ddpg} to train an end-to-end DRL policy in a realistic Unity simulation environments. Our DRL network consists of three branches: 1. elevation map branch (2D input), 2. IMU vector branch and 3. the 1D parameter input branch. The DRL network architecture is depicted in Fig. \ref{fig:network-architecture}.

We utilize a robot-centric elevation map ($E_t$) as the 2D input to the first branch. It is passed through a convolution layer and an attention module named CBAM\cite{CBAM} to encode the elevation changes in the map.  This attention module is capable of performing channel and spatial attention to infer an attention feature vector that reflects critical elevations. Further, the attention formulation denoises low-level features at the early stage and later gradually focuses on high-level semantics such as the critical elevation gradients. 

% \adarsh{are these benefits explained later on?}. 

The second branch uses an IMU vector ($V_{IMU}$) that includes 6-dimensional IMU values from the last $T$ time steps.  This branch incorporates a set of LSTM \cite{lstm} layers to encode long-term dependencies between time steps in the IMU data sequence.
% \adarsh{temporal features; what are these features? Vibrations at different time instants? Must be explicitly mentioned.} from the robot's recent IMU data.
The third branch takes several one dimensional inputs such as the heading elevation gradient vector ($\nabla h$), the robot's orientation \big(roll($\phi$) , pitch($\psi$)\big),
% (\adarsh{$\theta_{roll}$, $\theta_{pitch}$; conventionally $\phi$ and $\psi$ are used I think?}),
previous actions in terms of linear and angular velocities ($v_{t-1},\omega_{t-1}$ respectively), and the heading angle and the distance to the goal respectively ($\alpha_{goal} , d_{goal}$, respectively) as inputs. Here, the heading elevation gradient ($\nabla h$) is the gradient vector calculated from the elevation map's vector ($h$) along the robot's heading direction. 

Finally, the three branches are concatenated and passed through a set of fully connected layers to obtain the output actions ($v,\omega$). 

\subsubsection{Reward Structure}
To train the policy to perform stable and smooth
% \adarsh{spatially aware; this word is used for the first time here. So maybe remove it, or explain it in Introduction.} 
navigation, we define a set of rewards as follows: 

\begin{figure}[t]
      \centering
      \includegraphics[width=\columnwidth,height=3cm]{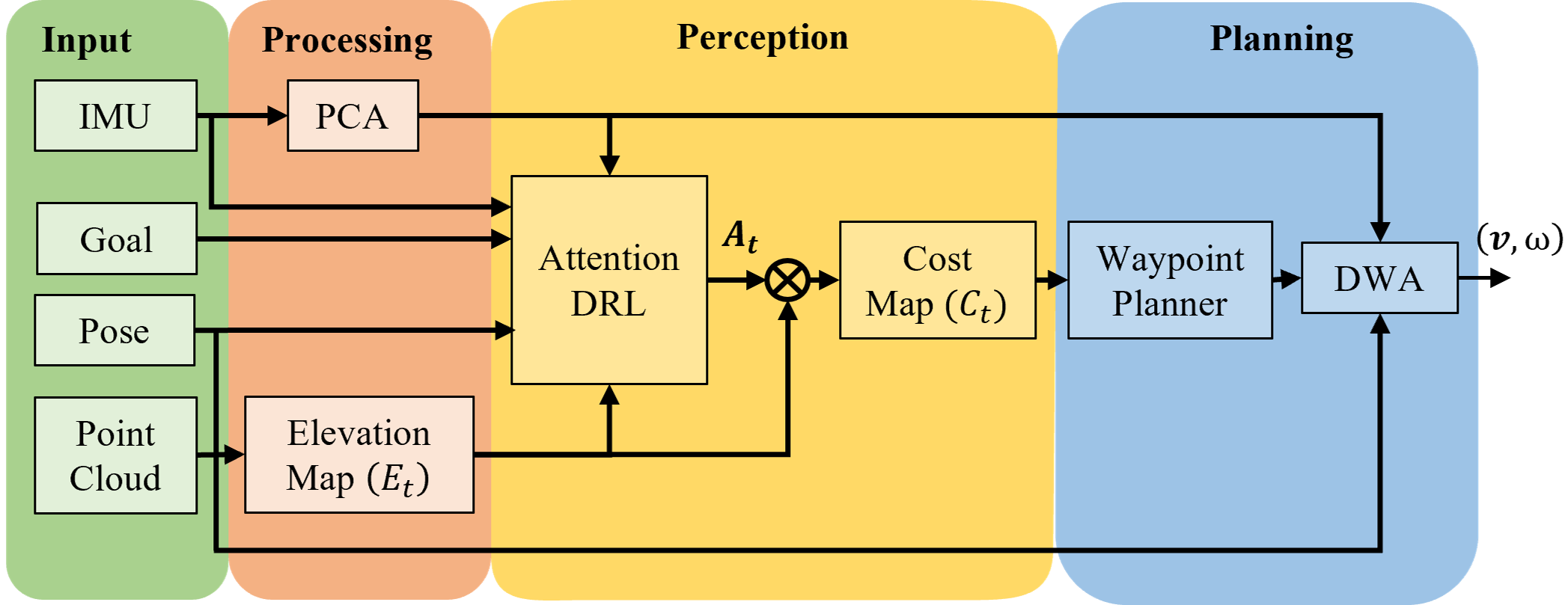}
      \caption {\small{\textbf{Our Overall System Architecture:} We propose a hybrid architecture to combine perception from the DRL module with our planning module. Instead of using actions from the end-to-end DRL network, we extract an intermediate output ($A_t$) from it to compute a navigation cost-map ($C_t$) to couple with our planner. This formulation displays comparable or better navigation performance in both simulated and real-world environments. Detailed analysis about the benefits of our method is presented in Section \ref{analysis}.  }}
      \label{fig:system-architecture}
      \vspace{-15pt}
\end{figure}

Goal reaching rewards, $R_{dist}$ and $R_{head}$ are defined as follows to penalize the robot's deviation from its goal.
\begin{equation}
    \begin{split}
        R_{distance} = - d_{goal}, \quad R_{heading} = - |\alpha_{goal}|.
    \end{split}
    \vspace{-10pt}
\end{equation}
$d_{goal}$ is the distance to the goal w.r.t. the robot and $\alpha_{goal}$ is the angle between the vector from the robot's position to the goal and the robot's heading direction. 

We introduce a novel reward, $R_{stable}$ to minimize high roll and pitch angle variations to maintain the robot's stability. 
\vspace{-1 pt}
\begin{equation}
        R_{stable} = - (|tanh(\phi)| + |tanh(\psi)|).
\vspace{-1 pt}
\end{equation}

To penalize navigating on bumpy regions (high elevation gradients), we define $R_{elev}$ as a weighted sum of the elevation gradients along the robot's heading direction. 
\vspace{-3 pt}
\begin{equation}
        R_{elev} = -\sum_{i=1}^{N_h} \nabla h_{i} e^{- ik_{elev}}. 
\end{equation}
$k_{elev}$ is a tunable parameter to adjust the rate of the exponential decay and weighs elevation gradients closer to the robot higher than the ones farther away.

Even though $E_t$ reflects elevation variations, it cannot identify fine grained details of the terrain that influence the vibrations experienced by the robot during navigation. To this end, we observe that dimension reduced IMU data can be used to measure a surface's bumpiness \cite{terrapnJ}. In particular, after applying PCA\cite{pca}, variances of the first two principal components of IMU data $(\sigma_{PC1},\sigma_{PC2})$ reflects a surface's level of vibration. Hence, we define our vibration reward as follows,

\vspace{-5 pt}
\begin{equation}
        R_{vibr} = -||\sigma_{PCA}||_2,
\end{equation}

where $\sigma_{PCA} = [\sigma_{PC1},\sigma_{PC2}]$. 

Finally, the total reward obtained for a given action is calculated as,
\begin{equation}
        R_{total} = \sum_{j}^{N_{rewards}}  \beta_{j} R_{j}, \,\,\, j \in \{dist,head,stable,elev,vibr\}
\end{equation}

where $N_{rewards}$ is the number of reward functions and $\beta_{j}$s are the weights corresponding to each component.

\begin{figure}[t]
      \centering
      \includegraphics[width=\columnwidth,height=4.5cm]{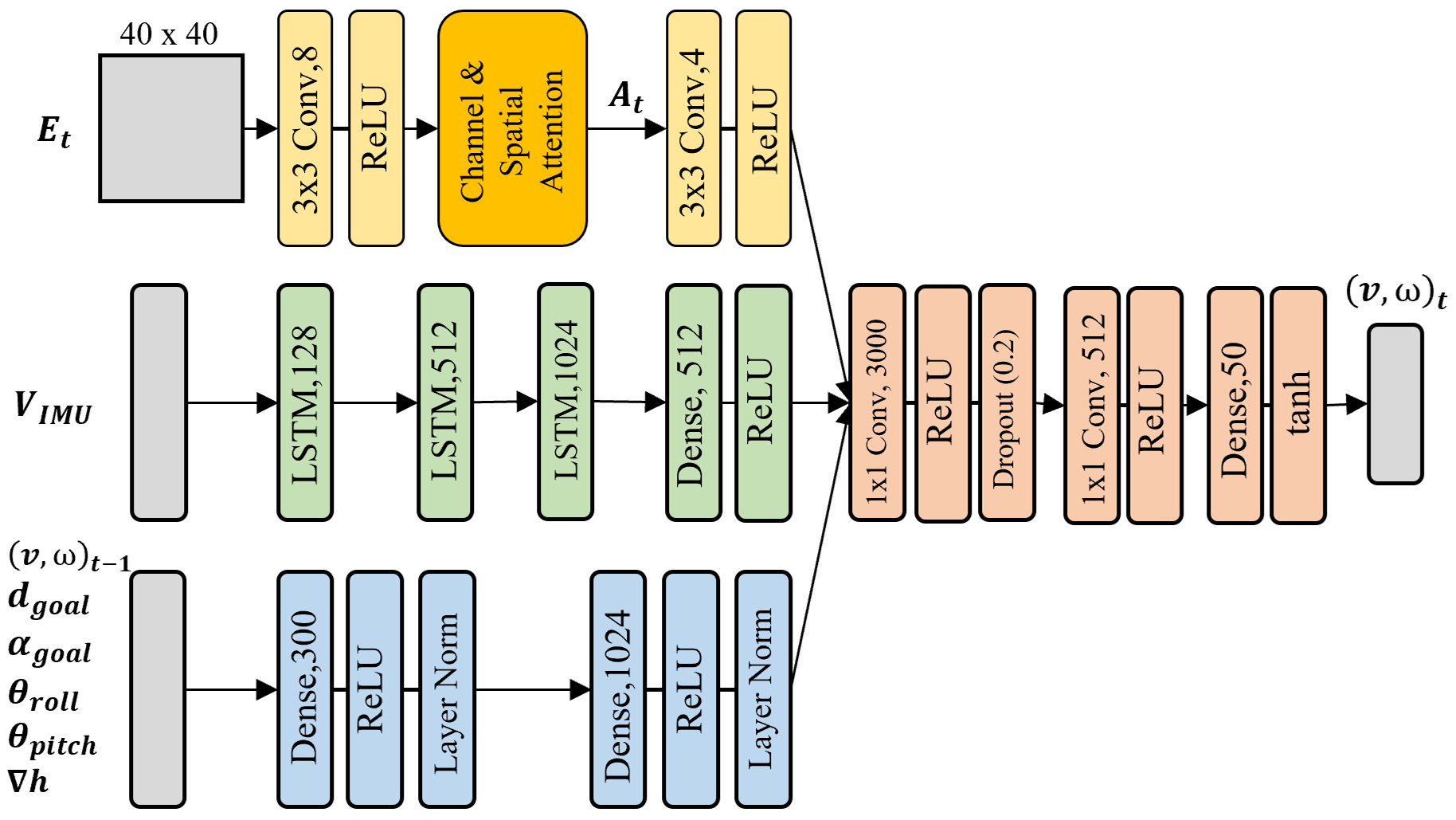}
      \caption {\small{\textbf{DRL Network Architecture:} We incorporate three input branches to feed observations into the DRL network. Elevation map $(E_t)$ branch consists of an attention module and CNN layers. Later we utilize output ($A_t$) from this attention module for perception. IMU vector branch consists of several LSTM layers to maximize the encoding of time varying IMU vector. The final branch incorporates fully connected layers to process the remaining one-dimensional observations.  }}
      \label{fig:network-architecture}
      \vspace{-15pt}
\end{figure}

Even though the fully trained Attention DRL network outputs the robot's velocities/actions for navigation, we do not use them for planning. Instead, we utilize an intermediate 
% \adarsh{output; maybe feature vector is a better word?} 
called the \textit{attention feature vector} $A_t$ (see Fig. \ref{fig:network-architecture}) to calculate a navigation cost map. 

\subsection{Navigation Cost-map Generation}
\begin{figure}[t]
      \centering
      \includegraphics[width=\columnwidth,height=3.2cm]{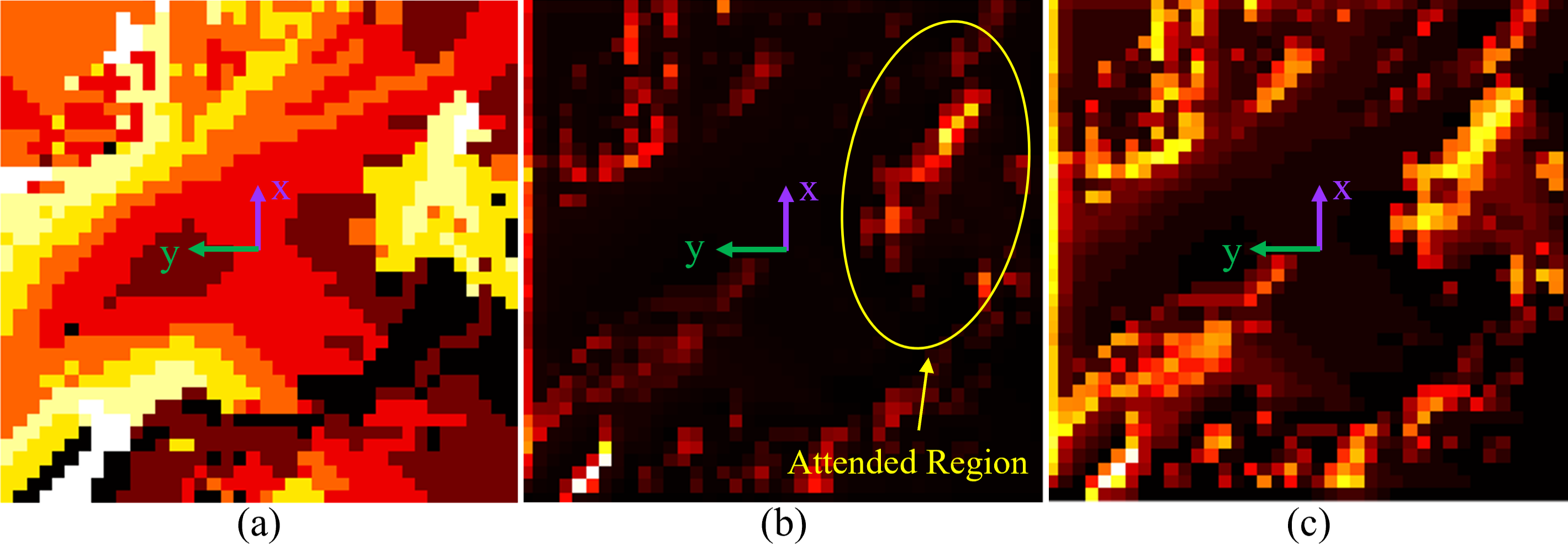}
      \caption {\small{\textbf{Navigation Cost-map Generation:} (a) Input elevation Map $(E_t)$; (b) Attention Feature Vector $(A_t)$; (c) Navigation Cost-map $(C_t)$. Dark colors (starting from black) indicate low values and bright colors (upto white) indicate high values in all three maps. We observe that the Attention Feature Vector $(A_t)$ only highlights critical elevation gradients towards the goal direction (i.e. top right at the moment). Further, the final costmap represents other critical elevations while focusing more on elevation gradients towards the goal.  }}
      \label{fig:cotmap_generation}
      \vspace{-15 pt}
\end{figure}

Once the \textit{attention feature vector} ($A_t$) is extracted, we combine it with the input elevation map ($E_t$) as follows to obtain the navigation cost map($ C_{t}$).
\vspace{-5 pt}
\begin{equation}
        C_{t} = A_t \odot E_t
\end{equation}
Here, all the maps $E_t, A_t$ and $C_t$ are $N \times N$ matrices. $\odot$ denotes the element-wise matrix multiplication. The maps extracted during the costmap generation process are presented in Fig. \ref{fig:cotmap_generation}.

\subsection{Waypoint Computation}
We adapt the minimum-cost waypoint calculation method from our previous work \cite{terp_arxiv} to obtain locally least-cost waypoints towards a given goal. This formulation guarantees that the resulting waypoints are along the stable and safest regions on a local cost-map (i.e. minimum cost regions in the navigation cost-map).

\subsection{DWA with Constrained Velocity Space}
We incorporate the well-known Dynamic Window Approach \cite{DWA} with adaptive velocity limits to follow the least-cost waypoints on the cost-map $C_t$. 

%DWA searches for the optimal velocity pair $(v,\omega)$ directly on a constrained velocity space by reducing the search space to a dynamic window, which consists of the reachable velocities within a short time interval.

Consider $V_s$ as the space with all possible velocities, and $V_a$ as the admissible velocity space containing the set of collision-free $(v, \omega)$ pairs. Further, let $V_d$ be the dynamic window space that includes only the dynamically feasible velocities within the next time interval $\Delta t$. Then, DWA formulates a restricted velocity space $V_r$ from the aforementioned velocity spaces on which an optimal velocity pair can be searched. Hence, $V_r$ can be denoted as $V_r = V_s \cap V_a \cap V_d$.

However, the velocities in $V_r$ could lead to robot flip-overs and high vibrations in uneven outdoor environments if the velocity and acceleration limits in DWA are too high. Alternatively, if these limits are too low, the robot may take a long time to reach its goal, or not be able to traverse highly sloped terrains. To overcome this issue, we introduce two novel constraints on the velocity search space to compute appropriate acceleration and velocity limits on uneven terrains.

Let $V_{el}$ be the set of velocities achievable without a robot flip-over and $(v_a, \omega_a)$ be the robot's current velocity. We adjust the linear velocity limit using the robot's current pitch angle and angular velocity limit using the roll angle. Hence, $V_{elev}$ is defined as,
\vspace{-5 pt}
\begin{equation}
\begin{split}
    V_{el} = \big\{ (v,\omega)| v \in [0, v_a + v_{el}^{pitch} ] , \\
    \omega \in  [\omega_a - \omega_{el}^{roll}, \omega_a + \omega_{el}^{roll} ] \big\},
\end{split}
\end{equation}

% \adarsh{$v_a, \omega_a$ are not defined} 
where $ v_{el}^{pitch} = \lambda_{el}tanh(\psi)$ only if $\psi \leq \psi_{lim}$; the linear velocity constraints are not applied otherwise. We avoid imposing constraints on $v$ if the robot's pitch angle is beyond the upper limit $\psi_{lim}$ to ensure that the velocity space contains large enough linear velocities to navigate in steep elevations. $\omega_{el}^{roll} =  \lambda_{el}|tanh(\phi)| $ and $v_{el}^{pitch}$ are orientation adaptive velocity limits with an adjustable parameter $\lambda_{el}$. 

Let $V_{vib}$ be the set of velocities achievable with low vibration (quantified from $||\sigma_{PCA}||_2$ ). We define $V_{vib}$ as,
\vspace{-5 pt}
\begin{equation}
\begin{split}
    V_{vib} = \big\{ (v,\omega)| v \in [0 , v_a -  v_{vib}^{lim}] , \\ \omega \in  [\omega_a -\omega_{vib}^{lim}, \omega_a + \omega_{vib}^{lim}] \big\},
\end{split}
\end{equation}

% \adarsh{$v_a, \omega_a$ are not defined} 
where $v_{vib}^{lim}= \omega_{vib}^{lim} = \lambda_{vib}||\sigma_{PCA}||_2$ with the tunable parameter $\lambda_{vib}$.  

Finally, the modified search space $V_{new}$ can be obtained as, $V_{new} = V_s \cap V_a \cap V_d \cap V_{el} \cap V_{vib}$. This improved search space is utilized to find the optimal $(v,\omega)$ pair by maximizing the following objective function,
\vspace{-5 pt}
\begin{equation}
    G(v,\omega) = \sigma \big(\alpha . head(v,\omega)+\beta. dist(v,\omega)+\gamma. vel(v,\omega)\big)
\end{equation}

Here, $head(v,\omega), dist(v,\omega)$ and  $vel(v,\omega)$ are cost functions defined in the DWA\cite{DWA} algorithm to provide higher values when the robot's heading is towards the goal, the distance to the goal is decreasing and the goal reaching velocity is high.

\section{Results and Analysis}
In this section, we present simulations and real-world robot implementations of our method. Further, we explain our evaluations and comparisons using different metrics.

\subsection{Implementation}

The end-to-end DRL network in Fig. \ref{fig:system-architecture} is implemented using Pytorch. We utilized a Unity based outdoor simulation, a Clearpath Husky robot model with a Velodyne VLP16 3D LiDAR, and ROS Melodic to train the network. The training and simulation are conducted in a workstation with an Intel Xeon 3.6 GHz processor and an Nvidia Titan GPU. 

The real world evaluations are performed using a real Husky robot, VLP16 LiDAR and a laptop with an Intel i9 CPU and an Nvidia RTX 2080 GPU. We use the Elevation Mapping ROS package \cite{Fankhauser2014RobotCentricElevationMapping} to generate the robot-centric elevation map ($\mathbf{E_t}$) of size $40 \times 40$.    

\subsection{Evaluations}

We compare our method's navigation performance with DWA\cite{DWA}, TERP\cite{terp_arxiv}, our end-to-end DRL network, and our method without the CBAM attention module. The following metrics are utilized to perform quantitative evaluations:

\no \textbf{Success Rate} - The percentage of successful goal reaching attempts out of the total number of experiments without any collisions or flip-overs.

\no \textbf{Avg. Vibration} - The average value of the PCA based vibration cost (i.e. $||\sigma_{PCA}||_2$ ) along a trajectory.

\no \textbf{Avg. Speed} - The robot's average velocity for a given path.

\no \textbf{Normalized Trajectory Length} - Navigation trajectory length normalized by the straight-line distance to the goal.

\subsection{Test Scenarios}
Let $EG_{max}$ denote the maximum elevation gain of an environment. We categorize elevated terrains as: Low ($EG_{max} \leq 1m$), Medium ($EG_{max} \sim 1-2m$) and High ($EG_{max} \geq 3m$). Then, we evaluate our method using the above metrics in three testing scenarios :\\
% \adarsh{This has to be quantified. Low, medium and high.}
\no \textbf{Scenario 1:} Low elevation  with multiple rough surfaces. \\
\no \textbf{Scenario 2:} Medium elevation  with multiple rough surfaces. \\
\no \textbf{Scenario 3:} High elevation  with one surface.

\begin{figure}[t]
      \centering
      \includegraphics[width=\columnwidth,height=6cm]{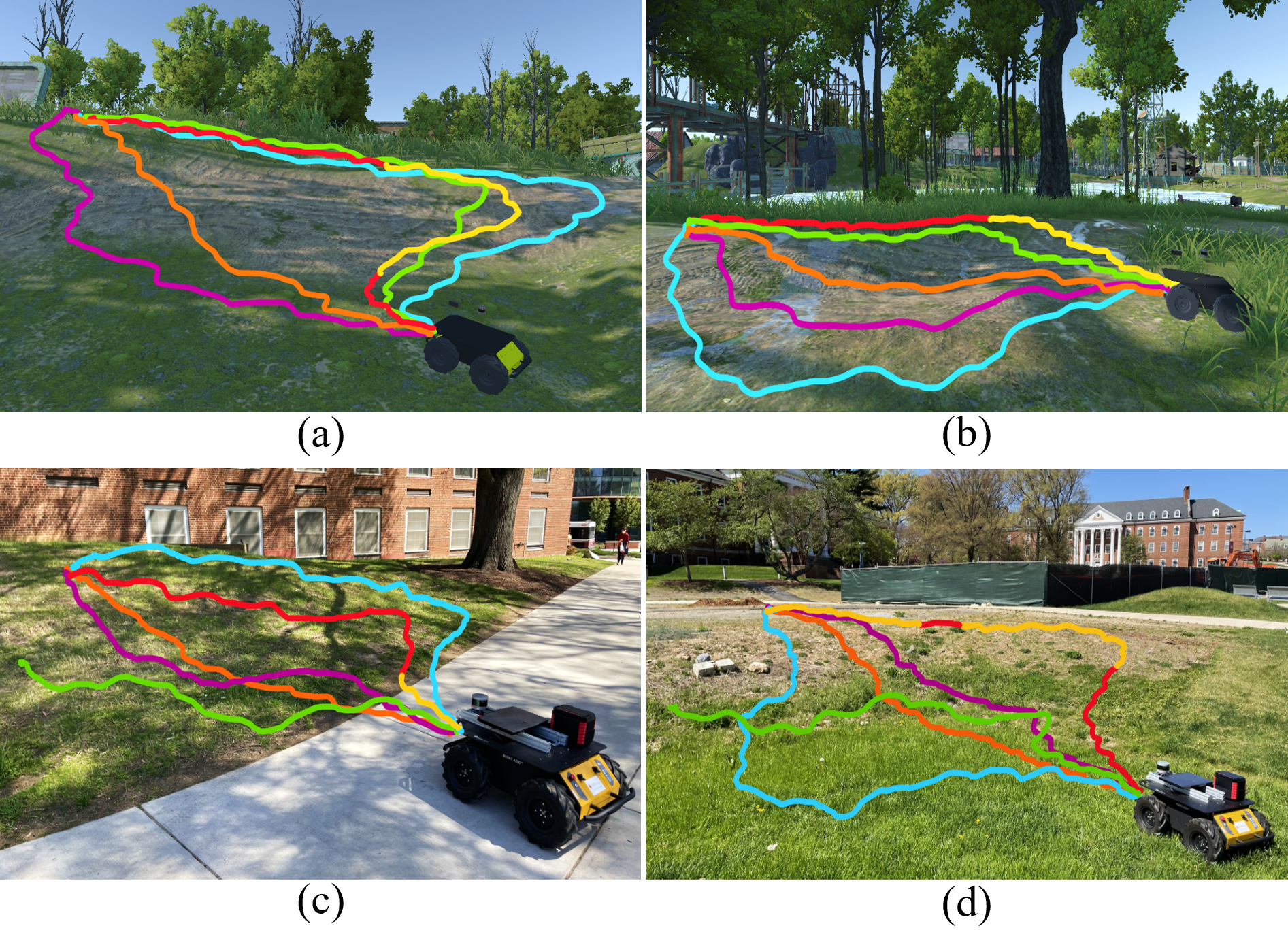}
      \caption {\small{\textbf{Navigation Comparisons:} Trajectories generated by  Ours(red and yellow for slow and fast speeds); Ours without attention(violet); End-to-end DRL (green); TERP(blue) and DWA(orange) when navigating in simulated and real-world uneven terrains. (a) Scenario 1 (simulated) (b) Scenario 2 (simulated) (c) Scenario 1(real) (d) Scenario 2(real). We observe that our method generates stable trajectories along less steep slopes with controlled velocities to minimize the risk of robot flip-overs. Further, it maintains comparable navigation performance in both simulated and real uneven terrains while end-to-end DRL method displays a significant performance degradation during sim-to-real transfer and does not reach the goal in the real world scenarios.   }}
      \label{fig:comparisons}
      \vspace{-15pt}
\end{figure}

\begin{table}
\resizebox{\columnwidth}{!}{%
\begin{tabular}{ |c |c |c |c |c |} 
\hline
\textbf{Metrics} & \textbf{Method} & \multicolumn{1}{|p{1cm}|}{\centering \textbf{Scenario} \\ \textbf{1}} & \multicolumn{1}{|p{1cm}|}{\centering \textbf{Scenario} \\ \textbf{2}} & \multicolumn{1}{|p{1cm}|}{\centering \textbf{Scenario} \\ \textbf{3}}\\ [0.5ex] 
\hline
\multirow{6}{*}{\rotatebox[origin=c]{0}{\makecell{\textbf{Success}\\\textbf{Rate (\%)}}}} 
 & DWA \cite{DWA} & 100 & 78 & 56  \\
 & TERP \cite{terp_arxiv} & 100 & 84 & 73   \\
 & End-to-end DRL & 98 & 72 & 61 \\
 & Ours without Attention & 99 & 75 & 67  \\
 
 & Ours with Attention & \textbf{100} & \textbf{89} & \textbf{78}  \\
\hline

\multirow{6}{*}{\rotatebox[origin=c]{0}{\makecell{\textbf{Avg.}\\\textbf{Vibration}}}}  
 & DWA \cite{DWA}  & 0.232 & 0.195 & 0.168  \\
 & TERP \cite{terp_arxiv} & 0.214 & 0.207 & 0.162  \\
 & End-to-end DRL & 0.168 & 0.181 & 0.127  \\
 & Ours without Attention & 0.116 & 0.103 & 0.091  \\
 
 & Ours with Attention & \textbf{0.097} & \textbf{0.093} & \textbf{0.079}  \\
\hline

\multirow{6}{*}{\rotatebox[origin=c]{0}{\makecell{\textbf{Avg.}\\\textbf{Speed}}}} 
 & DWA \cite{DWA}  & 0.684 & 0.671 & 0.659 \\
  & TERP \cite{terp_arxiv} & 0.637 & 0.644 & 0.628    \\
 & End-to-end DRL & 0.646 & 0.592 & 0.589  \\
 & Ours without Attention & 0.539 & 0.540 & 0.514  \\
 
 & Ours with Attention & \textbf{0.496} & \textbf{0.434} & \textbf{0.357}  \\
\hline

\multirow{6}{*}{\rotatebox[origin=c]{0}{\makecell{\textbf{Norm.}\\\textbf{Traj.}\\\textbf{Length}}}} 
 & DWA \cite{DWA}  & \textbf{1.008} & \textbf{1.124} & \textbf{1.116} \\
 & TERP \cite{terp_arxiv} & 1.112 & 1.236 & 1.307    \\
 & End-to-end DRL & 1.142 & 1.165 & 1.187  \\
 & Ours without Attention & 1.109 & 1.172 & 1.226  \\
 
 & Ours with Attention & 1.105 & 1.164 & 1.208  \\
\hline

\end{tabular}
}
\caption{\small{\textbf{Performance Comparisons:} Our method consistently maintains the highest success rate, minimum average vibration and the lowest average speed when navigating in complex uneven outdoor environments. However, our method result in relatively longer trajectories to avoid unstable regions while others generate shorter paths to reach the goal.}
}
\label{tab:comparison_table}
\vspace{-15pt}
\end{table}

\subsection{Analysis}
\label{analysis}

We evaluate our method's navigation performance qualitatively in Fig. \ref{fig:comparisons} and quantitatively in Table. \ref{tab:comparison_table}. We observe that all the methods perform reasonably well in terms of success rate in Scenario 1. However, our method maintains higher success rates even in steep elevations such as Scenario 2 and 3 (see Fig. \ref{fig:comparisons} and \ref{fig:cover-image}), while other methods show a significant decrease in the success rate. Further, the average vibration encountered by the robot is significantly less when navigating using our approach. The trajectories generated by DWA and TERP result in high vibrations since they do not consider robot vibration during planning. 

We observe that our method's average navigation velocity is significantly lower in Scenarios 1 and 2 than the other methods. Even the end-to-end DRL method navigates at a higher average speed than our approach. This indicates that the velocity constraints applied for high vibrations and elevations in our DWA formulation are capable of reducing the navigation velocities to maintain the stability and smoothness of the trajectories. 

We further notice that the trajectory lengths of our approach are generally between the trajectory lengths of DWA and TERP.  DWA generates trajectories to maximize the goal reaching cost (i.e. short and relatively straight trajectories) without considering any terrain properties. In contrast, TERP computes way points along the locally least-cost regions and navigates without any terrain adaptive velocity constraints on the planner (which could lead to reasonably longer paths). However, our method's terrain-aware velocity constraints minimize the use of high angular velocities to deviate from the goal during waypoint-to-waypoint navigation.

\textbf{End-to-end vs Ours:} We observe a significant performance degradation in the end-to-end DRL network when navigating on real outdoor terrains (see Figs.\ref{fig:cover-image} and \ref{fig:comparisons}). However, our hybrid formulation ensures that the navigation performance is comparable or better in both simulated and real uneven outdoor environments.

\section{Conclusions, Limitations and Future Work}

We present a novel sim-to-real formulation to utilize perception features encoded in a fully trained DRL policy for robot navigation in unstructured outdoor environments. We generate a navigation cost-map using an intermediate result from a DRL network and perform waypoint-to-waypoint navigation using DWA with adaptive velocity constraints to avoid robot flip-overs. We validate and compare our method's capabilities in both simulated and real-world unstructured terrains. 
Our formulation has a few limitations. The robot cannot avoid rough terrains though the navigation speed can be reduced adaptively in such scenarios. To this end, vision based strategies can be used to identify surface properties to avoid rough terrains.

\bibliographystyle{IEEEtran}
\bibliography{References}

\end{document}